# EvoNF: A Framework for Optimization of Fuzzy Inference Systems Using Neural Network Learning and Evolutionary Computation

Ajith Abraham, *Member, IEEE*
School of Business Systems, Monash University, Clayton, Victoria 3168, Australia. Email: ajith.abraham@ieee.org

*Abstract--* Several adaptation techniques have been investigated to optimize fuzzy inference systems. Neural network learning algorithms have been used to determine the parameters of fuzzy inference system. Such models are often called as integrated neuro-fuzzy models. In an integrated neuro-fuzzy model there is no guarantee that the neural network learning algorithm converges and the tuning of fuzzy inference system will be successful. Success of evolutionary search procedures for optimization of fuzzy inference system is well proven and established in many application areas. In this paper, we will explore how the optimization of fuzzy inference systems could be further improved using a meta-heuristic approach combining neural network learning and evolutionary computation. The proposed technique could be considered as a methodology to integrate neural networks, fuzzy inference systems and evolutionary search procedures. We present the theoretical frameworks and some experimental results to demonstrate the efficiency of the proposed technique.

*Index terms--* fuzzy systems, neural networks, evolutionary computation, hybrid system

## I. INTRODUCTION

Artificial neural networks and fuzzy inference systems are both very powerful soft computing tools for solving a problem without having to analyze the problem itself in detail [1]. Natural intelligence is a product of evolution. Therefore, by mimicking biological evolution, we could also simulate high-level intelligence. Evolutionary computation works by simulating a population of individuals, evaluating their performance, and evolving the population a number of times until the required solution is obtained. The drawbacks pertaining to neural networks and fuzzy inference systems seem complementary and evolutionary computation could be used to optimize the integration to produce the best possible synergetic behavior to form a single system. The integrated architecture share data structures and knowledge representations [5]. The parameters of the fuzzy inference system will be fine-tuned using evolutionary algorithms and neural network learning techniques. In such an integrated environment, learning occurs at two levels: evolutionary learning (global optimization) and a local search by conventional neural network algorithm (gradient descent) [2]. Evolutionary algorithms could also be used to determine the optimal learning parameters of the gradient descent technique. It is interesting to note that Takagi-Sugeno-type fuzzy systems are high performers (more accuracy) but often requires complicated learning procedures and are computationally expensive. On the other hand, Mamdani type fuzzy systems can be modeled using faster heuristics but with a compromise on the performance (high RMSE). Hence there is always a compromise between performance and computational time [3]. Most of the integrated neuro-fuzzy systems currently available are based on either Mamdani-type or Takagi-Sugeno-fuzzy inference system. Hence selection of a good inference system itself becomes complicated when the dimensionality and complexity of the input-output mapping increases. For the success of a neuro-fuzzy design, the user has to specify the shape and quantity of the membership function for each input/output variable, fuzzy operators, defuzzification method, fuzzy inference mechanism etc. The user also has to specify the rule base (except EFuNN and NEFCON) and the learning technique that will fine-tune the membership functions and other tunable parameters [1]. We are familiar with "trapped in local minima" whenever we refer to local search techniques. Since the neuro-fuzzy systems use gradient descent method, there is no guarantee that global optima would be obtained and the parameters are fine-tuned. Evolutionary design of fuzzy systems has been investigated by several researchers [4] [7]. Evolutionary algorithms are popular for obtaining a global optimal solution [4] but not often well in local searches. Integrating evolutionary computation (a global optimization technique) with a local search technique might help to explore the solution space more effectively.

In Section II we present the theoretical frameworks for the proposed Evolving Neuro Fuzzy (EvoNF) system followed by chromosome modeling and representation issues in Section III. In section IV we present the experimentation setup and some discussions and conclusions are provided towards the end.

## II. GENERIC ARCHITECTURE OF EvoNF

In this section, we define the architecture of EvoNF, which is an integrated computational framework to optimize fuzzy inference system using neural network learning technique and evolutionary computation. The proposed framework could adapt to Mamdani, Takagi-Sugeno or other fuzzy inference systems. The architecture and the evolving mechanism can be considered as general framework for adaptive fuzzy systems, that is a fuzzy inference system that can change their membership functions (quantity and shape), rule base (architecture), fuzzy operators and learning parameters according to different environments without human intervention. Solving multi-objective scientific and engineering problems is, generally, a very difficult goal. In these particular optimization problems, the objectives often conflict across a high-dimension problem space and may also require extensive computational resources. We propose a 5-tier evolutionary search procedure wherein the membership functions, rule base (architecture), fuzzy inference mechanism (T-norm and T-conorm operators), learning parameters and finally the type of inference system (Mamdani, Takagi-Sugeno etc.) are adapted according to the environment. Figure 1 illustrates the interaction of various evolutionary search procedures. Referring to Figure 1, for every fuzzy inference system, there exist a global search of learning algorithm parameter, inference mechanism, rule base and membership functions in an environment decided by the problem. Thus the evolution of the fuzzy inference system will evolve at the slowest time scale while the evolution of the quantity and type of membership functions will evolve at the fastest rate. The function of the other layers could be derived similarly [5].

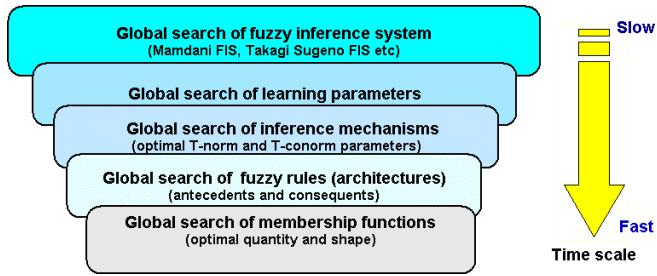

**Figure 1.** General computational framework for EvoNF

Hierarchy of the different adaptation layers (procedures) will rely on the prior knowledge. For example, if there is more prior knowledge about the architecture than the inference mechanism then it is better to implement the architecture at a higher level. If we know that a particular fuzzy inference system will suit best for the problem, we could also minimize the search space. For fine-tuning the fuzzy inference system all the node functions are to be parameterized.

**Parameterization of Membership Functions**

Fuzzy inference system is completely characterized by its Membership Function (MF). For example, a generalized bell MF is specified by three parameters (*p, q, r*) and is given by:

$$\text{Bell}(x, p, q, r) = \frac{1}{1 + \left|\frac{x - r}{p}\right|^{2q}}$$

Figures 2 shows the effects of changing *p, q* and *r* in a bell MF. Similar parameterization can be done with most of the other MF's [8].

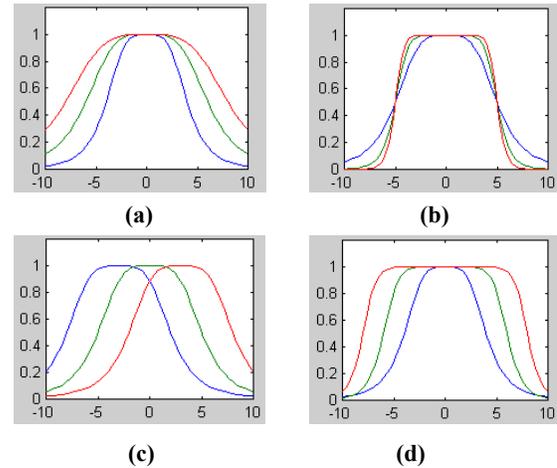

(a)  (b)

(c)  (d)

**Figure 2. (a)** Changing parameter *p* **(b)** changing parameter *q* **(c)** changing parameter *r* **(d)** changing *p* and *q*

**Parameterization of T-norm operators**

T-norm is a fuzzy intersection operator, which aggregates the intersection of two fuzzy sets A and B. The Schweizer and Sklar's T-norm operator can be expressed as [8] [9]:

$$T(a, b, p) = \left[\max\left\{0, (a^{-p} + b^{-p} - 1)\right\}\right]^{-\frac{1}{p}}$$

It is observed that

$$\lim_{p \to 0} T(a, b, p) = ab$$

$$\lim_{p \to \infty} T(a.b, p) = \min\{a, b\}$$

which correspond to two of the most frequently used T-norms in combining the membership values on the premise part of a fuzzy *if-then* rule.

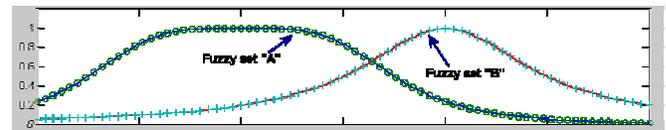

**Figure 3.** Bell MFs for fuzzy set *A* and *B*

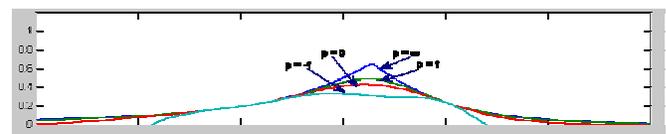

**Figure 4.** Effects of changing *p* of T-norm operator

To give a general idea of how the parameter *p* affects the T-norm operator, Figure 3 shows two bell shaped MFs *A* and

*B* and Figure 4 illustrates T-norm operator $T_{(a,b,p)}$ for different values of *p*.

## III. CHROMOSOME MODELING AND REPRESENTATION

The antecedent of a fuzzy rule defines a local region, while the consequent describes the behavior within the region via various constituents. Basically the antecedent part remains the same regardless of the inference system used. Different consequent constituents result in different fuzzy inference systems. For applying evolutionary algorithms, problem representation (chromosome) is very important as it directly affects the proposed algorithm. Referring to Figure 1 each layer (from fastest to slowest) of the hierarchical evolutionary search process has to be represented in a chromosome for successful modeling of EvoNF. A typical chromosome of the EvoNF would appear as shown in Figure 5 and the detailed modeling process is as follows.

**Layer 1:** The simplest way is to encode the number of membership functions per input variable and the parameters of the membership functions. Figure 7 depicts the chromosome representation of *n bell* membership functions specified by its parameters *p*, *q* and *r*. The optimal parameters of the membership functions located by the evolutionary algorithm will be later fine tuned by the neural network-learning algorithm. Similar strategy could be used for the output membership functions in the case of a Mamdani fuzzy inference system. Experts may be consulted to estimate the MF shape forming parameters to estimate the search space of the MF parameters.

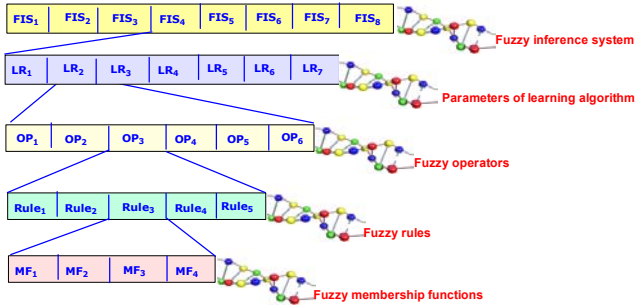

**Figure 5.** Chromosome structure of the EvoNF model

We used the angular coding method proposed by Cordón et al for representing the rule consequent parameters of the Takagi-Sugeno inference system [7]. Rather than directly coding the consequent parameters, the "transformed" parameters represent the direction of the tangent $\alpha_i = \arctan p_i$. The range for the parameters $\alpha_i$ is the interval $(-90^0, +90^0)$, such that the parameters $p_i$ can assume any real value. A single input Takagi-Sugeno system $Y = p_1 X + p_0$ defines a straight line. The real value $p_1$ is simply the gradient between this line and the *X*-axis. Parameter $p_0$ determines the offset of the straight line (intercept) along the *Y*-axis. Angular coding is advantageous, since the value of $p_0$ varies between different rules and it is difficult to use some fixed interval to exploit the search space. The procedure is illustrated in Figure 6.

**Layer 2.** This layer is responsible for the optimization of the rule base. This includes deciding the total number of rules, representation of the antecedent and consequent parts. The number of rules grow rapidly with an increasing number of variables and fuzzy sets. The simplest way is that each gene represents one rule, and "1" stands for a selected and "0" for a non-selected rule. Figure 8 displays such a chromosome structure representation. To represent a single rule a position dependent code with as many elements as the number of variables of the system is used. Each element is a binary string with a bit per fuzzy set in the fuzzy partition of the variable, meaning the absence or presence of the corresponding linguistic label in the rule. For a three input and one output variable, with fuzzy partitions composed of 3,2,2 fuzzy sets for input variables and 3 fuzzy sets for output variable, the fuzzy rule will have a representation as shown in Figure 9.

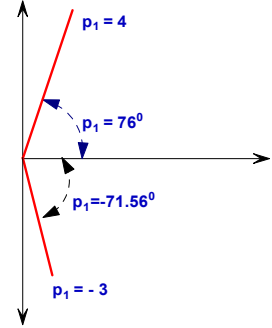

**Figure 6.** Angular coding technique of rule consequent parameters of Takagi Sugeno inference system

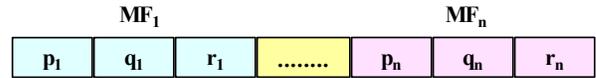

**Figure 7.** Chromosome representing *n m*embership functions for every input/output variable coding the parameters of a bell shape MF

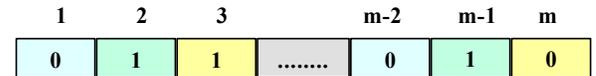

**Figure 8.** Chromosome representing the entire rule base consisting of m fuzzy rules

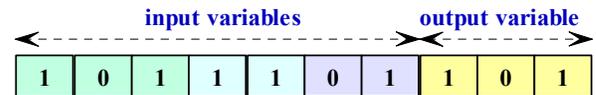

**Figure 9.** Chromosome representing an individual fuzzy rule (3 input variables and 1 output variable)

**Layer 3.** In this layer, a chromosome represents the different parameters of the T-norm and T-conorm operators. Real number representation is adequate to represent the fuzzy operator parameters. The parameters of the operators could be even fine- tuned using gradient descent techniques.

**Layer 4.** This layer is responsible for the selection of optimal learning parameters. Performance of the gradient descent algorithm directly depends on the learning rate according to the error surface. We used real number representation to represent the learning parameters. The optimal learning parameters decided by the evolutionary algorithm will be used to tune the membership functions and the inference mechanism.

**Layer 5.** This layer basically interacts with the environment and decides which fuzzy inference system (Mamdani type and its variants, Takagi-Sugeno type, Tsukamoto type etc.) will be the optimal according to the environment.

Once the chromosome representation, *C,* of the entire EvoNF model is done, the evolutionary search procedure could be initiated as follows:

1. *Generate an initial population of N numbers of C chromosomes. Evaluate the fitness of each chromosome depending on the problem.*
2. *Depending on the fitness and using suitable selection methods reproduce a number of children for each individual in the current generation.*
3. *Apply genetic operators to each child individual generated above and obtain the next generation.*
4. *Check whether the current model has achieved the required error rate or the specified number of generations has been reached. Go to Step 2.*
5. *End*

### IV. EXPERIMENTATION SETUP USING EVONF

We have applied the proposed technique to the three well known chaotic time series namely Mackey Glass [11], gas furnace [10] and waste water [12]. Fitness value is calculated based on the RMSE achieved on the test set. We have considered the best-evolved EvoNF model as the best individual of the last generation. We also explored three different learning methods:

**Type 1:** Evolutionary learning of membership functions, T-norm operator, rule base, consequent parameters and fine tuning of the membership functions using gradient descent method. The evolutionary algorithm further optimizes the learning rates of the gradient descent technique. This method could be considered as a meta-learning approach.

**Type 2:** Evolutionary learning of membership functions, T-norm operator, rule base and consequent parameters. No gradient descent learning is used. This is equivalent to pure evolutionary design of fuzzy inference systems.

**Type 3:** Evolutionary learning of membership functions, rule base and consequent parameters with fixed T-norm (min) operator. The MFs are fine-tuned using gradient descent method and the evolutionary algorithm is used to further optimize the learning rates. This experiment is to demonstrate how important is the tuning of fuzzy operators. For all the experiments, we reduced the search space by incorporating the following priori knowledge

- Takagi-Sugeno fuzzy inference system was selected
- The initial rule base was generated using a grid partitioning method and the rule base was further optimized using the evolutionary algorithm. This approach seems to work faster than building up the rule base from scratch.
- only Gaussian and Bell shaped MF's was used.

The genotypes were represented by real coding using floating-point numbers and the initial populations were randomly created based on the parameters shown in Table 1. We used a special mutation operator, which decreases the mutation rate as the algorithm greedily proceeds in the search space [7]. If the allelic value $x_i$ of the *i*-th gene ranges over the domain $a_i$ and $b_i$ the mutated gene $x_i'$ is drawn randomly uniformly from the interval $[a_i, b_i]$.

$$x_i' = \begin{cases} x_i + \Delta(t, b_i - x_i), if\ \omega = 0 \\ x_i + \Delta(t, x_i - a_i), if\ \omega = 1 \end{cases}$$

where $\omega$ represents an unbiased coin flip $p(\omega =0) = p(\omega =1) = 0.5$, and

$$\Delta(t, x) = x \left( 1 - \gamma^{\left(1 - \frac{t}{t_{\max}}\right)^b} \right)$$

defines the mutation step, where $\gamma$ is the random number from the interval $[0,1]$ and $t$ is the current generation and $t_{max}$ is the maximum number of generations. The function $\Delta$ computes a value in the range $[0,x]$ such that the probability of returning a number close to zero increases as the algorithm proceeds with the search. The parameter $b$ determines the impact of time on the probability distribution $\Delta$ over $[0,x]$. Large values of $b$ decrease the likelihood of large mutations in a small number of generations. The parameters mentioned in Table 1 were decided after a few trial and error approaches. Experiments were repeated 3 times for the three time series and the worst performance measures are reported. Table 2 summarizes the quantity of membership functions and the rule base before and after Type 1 learning. Test results showing the RMSE for the three time series are presented in Table 3. Learning convergence for the gas furnace, waster water and Mackey glass time series are plotted in Figures 10 - 12. For interest, empirical results of EvoNF was compared with ANFIS (Adaptive Neuro-Fuzzy Inference System)[8] implementing a Takagi-Sugeno fuzzy inference system and is depicted in Table 3.

TABLE 1. PARAMETERS USED FOR EVOLUTIONARY DESIGN OF NEURO-FUZZY SYSTEMS

| Population size | | | 30 | | | | | |
|---|---|---|---|---|---|---|---|---|
| Fuzzy inference system | | | Takagi Sugeno | | | | | |
| Rule antecedent membership functions | | | 2 - 4 membership functions per input, parameterized Gaussian | | | | | |
| Rule consequent parameters | | | angular coding | | | | | |
| T-norm operators | | | Parameterized Schweizer and Sklar's operator | | | | | |
| Learning rate | | | 0.05 – 0.20 | | | | | |
| Learning epochs | | | 100 epochs of gradient descent algorithm for all the 3 time series | | | | | |
| Ranked based selection | | | 0.50 | | | | | |
| Elitism | | | 5 % | | | | | |
| Starting mutation rate | | | 0.70 | | | | | |
| **Iterations** | | | | | | | | |
| **Mackey Glass** | Type 1 | 60 | **Gas Furnace** | Type 1 | 60 | **Waste Water** | Type 1 | 65 |
| | Type 2 | 90 | | Type 2 | 135 | | Type 2 | 180 |
| | Type 3 | 60 | | Type 3 | 60 | | Type 3 | 65 |

TABLE 2. COMPARISON OF MEMBERSHIP FUNCTIONS AND FUZZY RULES BEFORE AND AFTER LEARNING (TYPE 1)

| | Before learning | | | | | | After learning | | | | | |
|---|---|---|---|---|---|---|---|---|---|---|---|---|
| | MFs | | | | | | MFs | | | | | |
| | Input | | | | Output | No. of rules | Input | | | | Output | No. of rules |
| | $I_1$ | $I_2$ | $I_3$ | $I_4$ | $O_1$ | | $I_1$ | $I_2$ | $I_3$ | $I_4$ | $O_1$ | |
| Mackey Glass | 4 | 4 | 4 | 4 | linear | 256 | 3 | 3 | 4 | 3 | linear | 94 |
| Gas Furnace | 3 | 3 | - | - | linear | 9 | 3 | 4 | - | - | linear | 12 |
| Waste Water | 4 | 4 | 4 | 4 | linear | 256 | 4 | 3 | 4 | 3 | linear | 112 |

TABLE 3. COMPARISON OF EvoNF AND ANFIS

| | | Neuro-fuzzy | | | |
|---|---|---|---|---|---|
| Time series | Learning algorithm | EvoNF RMSE | | ANFIS RMSE | |
| | | Training | Test | Training | Test |
| Mackey Glass | Type 1 | 0.0007* | 0.0008* | | |
| | Type 2 | 0.0009 | 0.0009 | 0.0019 | 0.0018 |
| | Type 3 | 0.0009 | 0.0009 | | |
| Gas Furnace | Type 1 | 0.0093* | 0.0110* | | |
| | Type 2 | 0.0111 | 0.0154 | 0.0137 | 0.0570 |
| | Type 3 | 0.0101 | 0.0256 | | |
| Waste Water | Type 1 | 0.0150* | 0.0310* | | |
| | Type 2 | 0.0190 | 0.0342 | 0.0530 | 0.0810 |
| | Type 3 | 0.0180 | 0.0330 | | |

*Lowest RMSE values

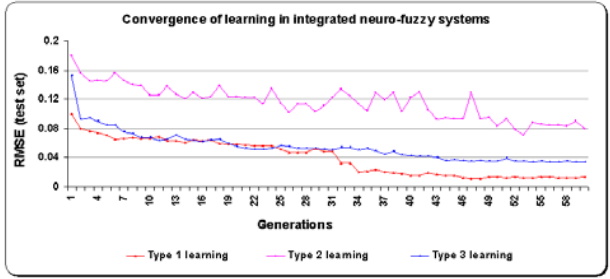

**Figure 10.** Gas furnace series learning

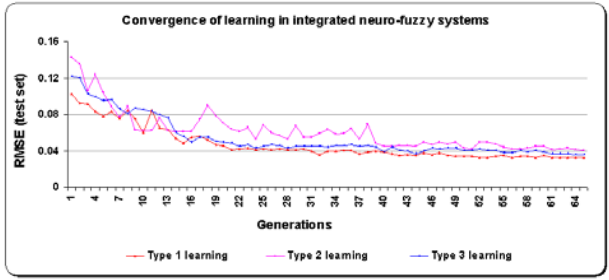

**Figure 11.** Waste water series learning

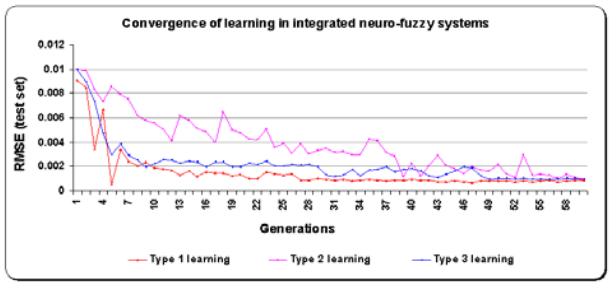

**Figure 12.** Mackey Glass series learning

V. DISCUSSIONS AND CONCLUSIONS

In this paper we have presented EvoNF, a computational framework for optimizing fuzzy inference systems using a 5-tier evolutionary search process. However, the real success in modeling such systems will directly depend on the genotype representation of the different layers. All prior knowledge available about the problem domain / system design are to be encoded into the system to minimize the search space by the evolutionary algorithms. We have explored the three different learning mechanisms to optimize a Takagi-Sugeno fuzzy inference system. As evident from Table 3, in terms of RMSE, the EvoNF model could outperform the conventional design of neuro-fuzzy systems using deterministic techniques. For all the three time series considered, EvoNF gave the best results on training and test sets. Our experiments using the three different learning strategies also reveal the importance of fine-tuning the global search method using a local search method. Type 2 learning method took longer time for the convergence and the final RMSE values obtained were higher than Type 1 approach. The empirical results obtained from Type 1 and Type 3 learning clearly illustrates the role of fuzzy operator tuning. Figure 13 illustrates the comparison of EvoNF model with different integrated neuro-fuzzy models and an artificial neural network (ANN) trained using backpropagation algorithm for predicting the Mackey Glass time series [3].

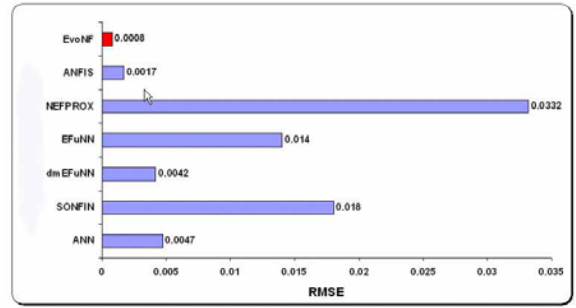

**Figure 13.** Comparison of EvoNF, neuro-fuzzy models and ANN

Hierarchical evolutionary search processes attract considerable computational effort. Fortunately, evolutionary algorithms work with a population of independent solutions, which makes it easy to distribute the computational load among several processors using parallel algorithms. Hence, for complicated problems, parallel evolutionary algorithms might prove to be very useful [6].